\pdfoutput=1

\documentclass[11pt]{article}

\usepackage[]{acl}

\usepackage{times}
\usepackage{latexsym}

\usepackage[T1]{fontenc}

\usepackage[utf8]{inputenc}

\usepackage{microtype}

\usepackage{inconsolata}

\usepackage{graphicx}
\usepackage{amsmath}
\usepackage{multirow}
\usepackage{tabularx}
\usepackage{booktabs}
\usepackage[most]{tcolorbox}

\title{Is Agentic RAG worth it? An experimental comparison of RAG approaches}

\author{\\
 \textbf{Pietro Ferrazzi\textsuperscript{1,2}},
 \textbf{Milica Cvjeticanin\textsuperscript{3}},
 \textbf{Alessio Piraccini\textsuperscript{4}},
 \textbf{Davide Giannuzzi\textsuperscript{5}}
\\
 \textsuperscript{1}Fondazione Bruno Kessler, Trento, Italy\\
 \textsuperscript{2}University of Padova, Italy \\
 \textsuperscript{3}Cargill Geneve, Switzerland\\
 \textsuperscript{4}Alkemy, Milan, Italy\\
 \textsuperscript{5}Komebi Studio, Milan, Italy\\
  \small{
    \textbf{Correspondence:} \href{mailto:pferrazzi [at] fbk [dot] eu}{pferrazzi [at] fbk [dot] eu}
  }
}

\begin{document}
\maketitle
\begin{abstract}
Retrieval-Augmented Generation (RAG) systems are usually defined by the combination of a generator and a retrieval component that extracts textual context from a knowledge base to answer user queries. However, such basic implementations exhibit several limitations, including noisy or suboptimal retrieval, misuse of retrieval for out-of-scope queries, weak query–document matching, and variability or cost associated with the generator. These shortcomings have motivated the development of "Enhanced" RAG, where dedicated modules are introduced to address specific weaknesses in the workflow.
More recently, the growing self-reflective capabilities of Large Language Models (LLMs) have enabled a new paradigm, often referred to as "Agentic" RAG. In this approach, an LLM orchestrates the entire process, deciding which actions to perform, when to perform them, and whether to iterate. Despite the rapid adoption of both paradigms, it remains unclear which approach is preferable under which conditions.
In this work, we conduct an empirically driven evaluation of "Enhanced" and "Agentic" RAG across multiple scenarios and dimensions. Our results provide practical insights into the trade-offs between the two paradigms, offering guidance on selecting the most effective RAG design for real-world applications, considering both performance and costs.

\end{abstract}


\begin{figure*}[t]
  \includegraphics[width=\textwidth]{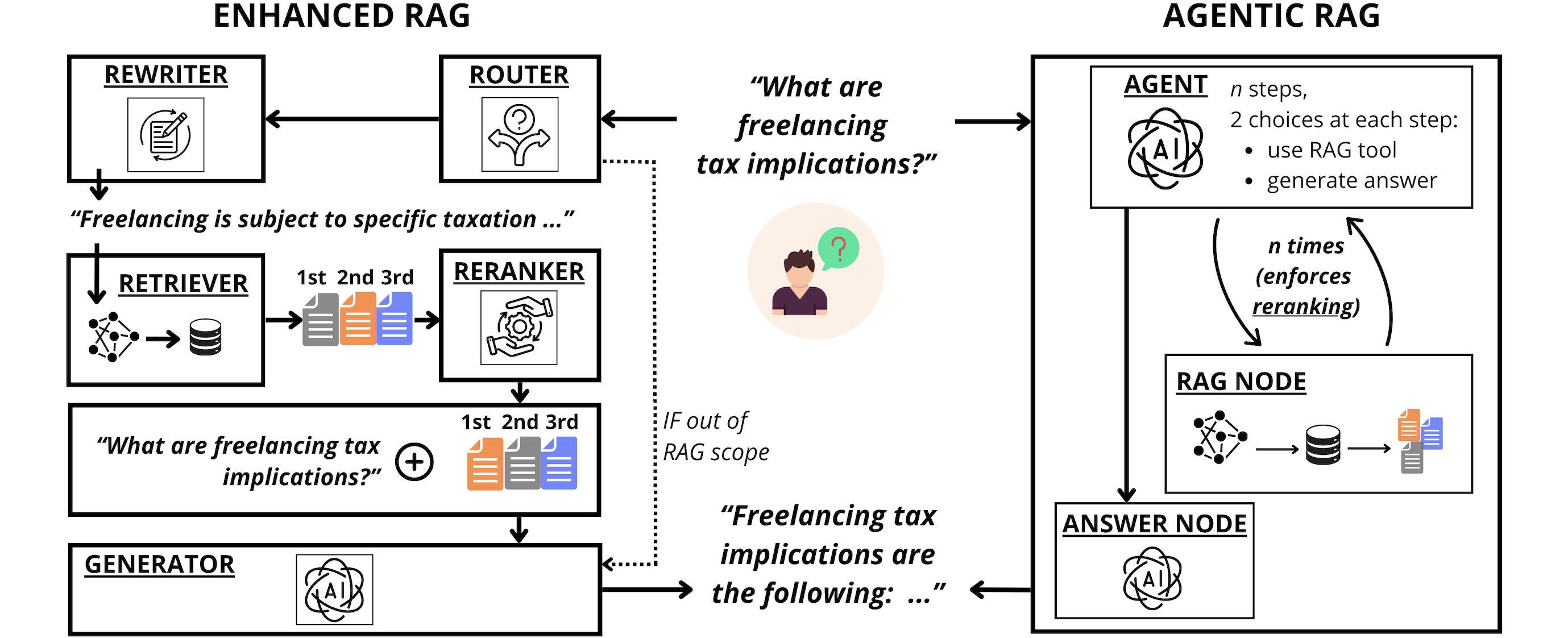}
\caption{\textbf{Left} — \textbf{Enhanced RAG}. The system is composed by a sequence of modules, each responsible for improving a specific stage of the RAG pipeline. A router determines whether a query should trigger retrieval; a rewriter reformulates the query; a retriever selects candidate chunks from the knowledge base; and a reranker orders the retrieved context before passing it to the generator. The workflow is fixed: information flows through predefined blocks intended to mitigate known weaknesses of \textbf{naïve RAG} systems (defined by the simple composition of the \textit{retriever} and \textit{generator} blocks).
\textbf{Right} — \textbf{Agentic RAG}. The LLM acts as an agent that orchestrates the entire process. At each step, it can choose to call a RAG tool or proceed to answer generation. Retrieval and context refinement can be repeated, as the agent autonomously selects operations based on the evolving state of the task.}
\label{fig:fig1}
\end{figure*}

\section{Introduction}
Retrieval-Augmented Generation (RAG) has evolved from a research concept \citep{NEURIPS2020_6b493230} into a core component of production-grade language systems, playing a central role in driving digital transformation across organizations \citep{ARSLAN20243781}.
This shift has fostered attention 
by both the research community \citep{wang-etal-2024-searching, 10.1145/3637528.3671470} and industry, 
with cloud providers offering their own RAG solutions for applications like enterprise QA, search assistants, internal knowledge bots \citep{ibm_watsonx_rag, aws_bedrock, azure_ai_search_rag}.
Since the first initial definitions, 
RAG workflows have been expanded to the so-called \textbf{Enhanced RAG} (Figure~\ref{fig:fig1}, left).
Such systems add to the retrieval and generation blocks components that perform further refinement.
Recently, LLMs' increasing self-reflective capabilities have enabled a shift towards \textbf{Agentic RAG} (Figure~\ref{fig:fig1}, right), where the LLM acts as an orchestrator, deciding which actions to perform, utilizing different tools for different purposes. Such systems are no longer fixed pipelines, but rather iterative loops guided by the model itself. 
Although initial work on identifying theoretical distinctions between Enhanced and Agentic RAG systems has been proposed \citep{neha2025traditional}, it remains unclear what the performance differences are between the two systems.
To this end, we aim to extract actionable insights for practitioners by analyzing performances and costs.
Our research question can be stated as follow:
\newtcolorbox{rqbox}{
  colback=blue!1,
  colframe=black,
  boxrule=0.8pt,
  arc=6pt,
  left=6pt,
  right=6pt,
  top=4pt,
  bottom=4pt,
  width=\linewidth,
  enhanced
}
\begin{rqbox}
\textcolor{blue}{When designing RAG systems, should practitioners adopt Agentic architectures or more traditional (Enhanced) pipelines?}
\end{rqbox}
Our \textbf{first contribution} consists of an experiment-driven comparison of the two paradigms in four dimensions relevant to production environments (Table~\ref{tab:summary_of_eval_dimensions}).
Our \textbf{second contribution} consists of a detailed analysis of costs and computational time required by the two systems under several scenarios. 
Finally, we propose a practical summary of our findings, aiming to support informed architectural choices in real-world RAG deployments\footnote{As per track requirements: preprint at \url{https://arxiv.org/abs/2601.07711}, Industry Day at LREC2026}.



\begin{table*}[]
\small
\centering
\renewcommand{\arraystretch}{1.2}
\begin{tabular}{|
    >{\centering\arraybackslash}m{4.2cm}|
    >{\centering\arraybackslash}m{4.2cm}|
    >{\centering\arraybackslash}m{2.8cm}|
    >{\centering\arraybackslash}m{2.9cm}|
}
\hline
\textbf{Naïve RAG shortcoming}  &
\textbf{What we evaluate}&
\multicolumn{2}{c|}{\textbf{Implementations}} \\
&  & \textbf{Enhanced} & \textbf{Agentic} \\  \hline
Retrieval is performed even for queries that do not require it & Accuracy of RAG usage for in-scope and out-of-scope queries & Semantic routing system    & Agent decides whether to do retrieval or not    \\ \hline
Queries and documents in the KB differ in format or semantics, causing weak retrieval   & Impact of query rewriting techniques        & Hyde-based query rewriting & Agent rewrites query as it wishes                \\ \hline
Noisy or suboptimal retrieval            & Impact of retrieved document list refinement techniques        & Encoder-based re-ranker    & Agent can redo retrieval multiple times  \\ \hline
The underlying LLM is too weak   /  too slow / is too   expensive & Impact of selecting more / less powerful models   & Test different LLMs & Test different LLMs     \\  \hline            
\end{tabular}
\caption{\label{tab:summary_of_eval_dimensions}
    Summary of the evaluation dimensions we select. For each shortcoming in Naïve RAG, we define an evaluation dimension ("What we evaluate") and an implementation to test how Enhanced and Agentic RAG overcome such a limitation. }
\end{table*}
\normalsize

\section{Related work}

\paragraph{RAG}
The concept of RAG, first introduced by \citet{NEURIPS2020_6b493230}, has undergone intensive research. Comprehensive overviews are presented by \citet{gao2023retrievalaugmented, 10.1145/3637528.3671470, wang-etal-2024-searching}. 
As large language models (LLMs) have acquired the capacity to multi-step reasoning and reflection, their consistency has enabled a paradigm shift toward Agentic RAG solutions \citep{10.5555/3666122.3666499, 10.5555/3666122.3668141}. 
An overview of how to combine the reasoning capabilities of LLMs with RAG-like structures is presented by \citet{li2025agenticragdeepreasoning}.
While the definition of the properties that characterize AI agents has evolved \citep{masterman2024landscapeemergingaiagent}, this emerging research direction has not yet been comprehensively categorized within a unified taxonomy, with initial attempts made by \citet{singh2025agenticretrievalaugmentedgenerationsurvey, 202502.0406}.

\textcolor{black}{\paragraph{RAG in enterprise}
Industry reports consistently identify knowledge-grounded applications—such as question answering over proprietary data, enterprise search, and document understanding—as among the highest-value use cases of generative AI. Reports by \citet{mckinsey2023genai}, and \citet{deloitte2023genai} emphasize the importance of connecting language models to internal data sources to improve reliability and business impact. 
Examples of such are use cases where large databases of user-technician interactions on specific issues are leveraged to provide answers to new users' requests; applications to query the all set of internal resources by companies \citep{flamel}. Notably, the workshop on Generative AI and RAG Systems for Enterprise at CIKM'24 \citep{10.1145/3627673.3680117} collects a series of applications in enterprise settings that motivate our focus on RAG itself.}

\paragraph{Terms definitions}
%
\textbf{Naïve Rag} \citep{gao2023retrievalaugmented} is the simplest instantiation of the RAG paradigm, 
where a \textit{retrieval} step extracts a fixed number of documents to be combined with the query and passed to an LLM for the answer \textit{generation} step.
According to the taxonomy outlined in \citet{huang2024surveyretrievalaugmentedtextgeneration}, \textbf{Enhanced RAG} refers to any Naïve RAG pipeline augmented with additional steps designed to improve its. \textcolor{black}{Relevant examples are CRAG \citep{DBLP:journals/corr/abs-2401-15884}, Self-RAG \citep{asai2024selfrag}, RaCoT \citep{Cai_Zhang_Fan_Yang_Wang_2026}, and INSIGHT-RAG \citep{chen2026insightrag}, which all focus on enhancing the performances of basic pipelines.}
\textbf{Agentic RAG} \citep{yao2023react, li2025searcho1agenticsearchenhancedlarge, alzubi2025opendeepsearchdemocratizing} is a system in which the LLM assumes control over the workflow, being able to dynamically decide to perform actions. 

\paragraph{The need for empirical comparison}
A preliminary effort for experimental comparisons between Agentic and Enhanced RAG is presented by \citet{neha2025traditional}, who propose a set of definitions and evaluation dimensions but stop short of conducting a full empirical study. Others \citep{xi2025infodeepseek, NEURIPS2024_1435d2d0, Chen_Lin_Han_Sun_2024} limit the benchmarking to one of the two settings.

\paragraph{Agents design and implementation}
Several open-source frameworks have emerged to support the development of Agentic RAG systems (Table~\ref{tab:agent_imp_frameworks}), reflecting the rapid growth of this area, ranging from minimalist designs to rich abstractions.

\section{Evaluation} \label{sec:evaluation}

The evaluation of RAG systems has been frequently decomposed into assessments of individual sub-components \citep{es-etal-2024-ragas, 10.1145/3399579.3399867}, each corresponding to a dimension that influences overall effectiveness. We design our evaluation following the same approach. First, we identify a list of limitations of Naïve RAG pipelines based on the work done by \citet{huang2024surveyretrievalaugmentedtextgeneration}, which we formulate in Table~\ref{tab:summary_of_eval_dimensions}. Then, we construct an experimental setting to compare how two implementations of Enhanced and Agentic RAG systems\footnote{We build Agentic Rag on PocketFlow \citep{pockeyflow_repo}} address each of them.
We focus on single-tool Agentic RAG systems, where the agent can only decide to invoke retrieval or produce a final answer. This design choice keeps its functional scope comparable to Enhanced RAG pipelines: multi-tool agents introduce additional capabilities (e.g., planning, external APIs) that would confound the comparison.
In the following sections, for each of the four identified dimensions, we \textit{i)} define it, \textit{ii)} detail our implementation choices, \textit{iii)} present the evaluation setting, \textit{iv)} define the evaluation metrics.

\subsection{Evaluation Datasets}
To conduct our experiments, we require datasets that are representative of common RAG applications, consisting of queries paired with a knowledge base. Following the taxonomy proposed by \citet{ARSLAN20243781}, which categorizes RAG use cases by application area, we focus on the most prominent natural language based scenarios. Specifically, we consider the two major categories: \textit{i)} Question Answering (QA), where RAG is utilized to ground answers in factual knowledge, and \textit{ii)} Information Retrieval and Extraction (IR/E), where RAG is intended as a tool to get knowledge from data through natural language queries.\\
For QA, we selected FIQA \citep{25} and NQ \citep{20} in the version released by \citet{beir}. For IR/E, we used FEVER \citep{35} and CQADupStack-English \citep{16}. Each dataset is chosen to represents a different real-world scenario and task type, as described in Table~\ref{tab:four_datasets}.

\begin{table*}
  \centering
\renewcommand{\arraystretch}{1.2}
  \small
  \begin{tabular}{
  |m{0.6cm}<{\centering\arraybackslash}
  |m{1.2cm}<{\centering\arraybackslash}
  |m{1.0cm}<{\centering\arraybackslash}
  |m{1cm}<{\centering\arraybackslash}
  |m{1.1cm}<{\centering\arraybackslash}
  |m{0.4cm}<{\centering\arraybackslash}
  |m{7.6cm}<{\arraybackslash}|}
    \hline
    \textbf{Task} & \textbf{Domain} & \textbf{Dataset} &
    \textbf{\#Query} & \textbf{\#Doc} & \textbf{Avg D/Q} & \textbf{Task description}\\
    \textbf{} & \textbf{} & \textbf{} & \textbf{} & \textbf{} & \textbf{} & \textbf{} \\
    \hline

    QA & General & NQ&
    3,452 & 2,681,468 & 1.2 &  Broad QA use cases, where users can ask any type of question to be answered via knowledge retrieval  \\
    \hline

    QA & Finance & FiQA &
    648 & 57,638 & 2.6 & Domain-specific queries to be answered by grounding responses on expert knowledge \\
    \hline

    IE/R & Grammar forum & CQAD-EN&
    1,570 & 40,221 & 1.4 & Find previously resolved blog posts that address the same question posed by the user, providing user-friendly summary \\
    \hline

    IE/R & Wikipedia & FEVER &
    6,666 & 5,416,568 & 1.2 & Seek evidence for or against the user statement ("claim verification") by finding documents and returning a final assessment with a summary of the references \\
    \hline
  \end{tabular}

  \caption{\label{tab:four_datasets}
    Overview of the four selected datasets used for the experimental settings.
    For both Question Answering (QA) and Information Extraction and Retrieval (IE/R),
    2 datasets are selected. Each query has a labelled list of relevant and irrelevant documents.
    Avg.\ D/Q indicates the average number of relevant documents per query.
  }
\end{table*}
\normalsize


\subsection{User Intent Handling}
\paragraph{Definition}
We refer to user intent handling as the need of determining whether a certain query requires the usage of retrieval or not.
While prior surveys on RAG \cite{huang2024surveyretrievalaugmentedtextgeneration, gao2023retrievalaugmented, 10.1145/3637528.3671470} do not explicitly address nor mention it, \citet{wang-etal-2024-searching}  highlight its importance by proposing a dedicated classifier for this task. We argue that intent detection is crucial in real-world RAG systems, as it prevents unnecessary or inappropriate retrieval calls. 
\textcolor{black}{When a query is classified as out-of-scope, the subsequent system behavior is application-dependent (e.g., fallback responses, refusal, or parametric answering), and is therefore beyond the scope of this work, which focuses solely on the routing decision itself.}

\paragraph{Enhanced Implementation}
We implement an Enhanced RAG routing system using the semantic-router framework 
\citep{semantic_router}
A router is defined by two sets of example queries, labelled as \textit{valid} and \textit{invalid} respectively. 
At inference time, the user query is compared to these groups
and it is classified as \textit{valid} or \textit{invalid} accordingly. The system uses RAG to answer valid queries and avoids it for invalid ones. 
For our experiments, we utilize OpenAI’s \texttt{text-embedding-3-small} as embedder. 
More details on the structure of the routing system are reported in Appendix~\ref{app:router}.

\paragraph{Agentic Implementation}
An Agentic RAG system embeds the ability to discriminate between queries that require retrieval by design. When a query is received, the agent can freely decide whether to utilise the RAG node or answer directly. 

\paragraph{Experimental setting}
We tested performances on a dataset composed by an equal number of valid and invalid queries ($500$ for each of the four datasets). 
We selected the valid queries from the train splits of each dataset, while we generated the invalid ones prompting \textit{gpt-4o} via 5-shot. We validate the invalid queries generation by calculating their average similarity with the valid ones (Appendix~\ref{app:invlaid_gen}). We make publicly available the datasets of valid and invalid queries\footnote {\url{https://huggingface.co/datasets/anonymousubmission/user-intent-handling}}.
We excluded the NQ dataset from this evaluation stage as it handles by design any type of query, preventing the definition of \textit{invalid} ones.

\paragraph{Evaluation metric}
To evaluate if systems correctly handle queries, we utilized  F1 score and recall, to take into account performances on both the valid and invalid classes.

\paragraph{Results}
Table~\ref{tab:routing} reports the results for user intent handling. We found that Agentic slightly overperforms Enhanced settings in the FIQA and CDQADupStack-EN tasks. In the case of FEVER, the former underperforms the latter by a margin ($-28.8$ F1 points), due to a very low recall ($49.3$). This low recall stems from the system often using retrieval even in cases it should not. We attribute these results to the first two datasets having a very clear domain definition (finance, English grammar), whereas the FEVER task is much less restrictive by design, as it aims to verify user queries on factual information, which makes it harder for the agent to understand what requests are "valid". We report the prompts in Appendix \ref{app:prompt}.

\begin{table}[]
\centering
\small
\renewcommand{\arraystretch}{1.2}
\begin{tabular}{|l|rr|rr|rr|}
         \hline
& \multicolumn{2}{c|}{\textbf{QA}} & \multicolumn{4}{c|}{\textbf{IR/E}}  \\
         \hline
   \textbf{Setting}      & \multicolumn{2}{c|}{\textbf{FIQA}} & \multicolumn{2}{c|}{\textbf{FEVER}} & \multicolumn{2}{c|}{\textbf{CQA-EN}} \\
         \hline
         &   rec  &  F1    &  rec   &  F1     &   rec    &   F1       \\
         \hline
naïve   & 100&66.7  & 100& \underline{66.7}   &100 & 66.7     \\
\hline
enhanced & 95.1  & \underline{95.7} &  84.4 & \textbf{87.9} &   94.7    & \underline{96.6}     \\
\hline
agentic  & 97.7   & \textbf{98.8}  &49.3    & 64.6   &  100      & \textbf{99.8}   \\
\hline
\end{tabular}
\caption{\textbf{User intent handling peformances} (recall and F1) on $500$ valid and invalid queries per dataset. The baseline is a Naïve RAG, where retrieval is performed for each user query. The Enhanced settings are based on the semantic router approach, while the Agent autonomously decided whether to use the RAG tool.}
\label{tab:routing}
\end{table}
\normalsize

\subsection{Query Rewriting}

\begin{table}[]
\centering
\renewcommand{\arraystretch}{1.2}
\small
\begin{tabular}{|l|c|c|c|c|c|}\hline
& \multicolumn{2}{c|}{\textbf{QA}} & \multicolumn{2}{c|}{\textbf{IR/E}} &  \\
\hline
               \textbf{Setting}                      & \textbf{FIQA} &  \textbf{NQ}   & \textbf{FEVER} & \textbf{CQAD} & \textbf{AVG}    \\
\hline
naïve & 45.3   & 43.7 & 66.2  & 45.8           & 50.3  \\
\hline
enhanced             & 43.5 & 43.9 & 81.1   & 42.8         & \underline{52.8} \\
\hline
agentic            & 43.2 &51.7& 83.1 & 44.3         & \textbf{55.6}  \\
\hline
\end{tabular}
\caption{\textbf{Query rewriting performances} in terms of NDCG@10. Naïve RAG represents the baseline where the user query is directly embedded without rewriting.}
\label{tab:rewriting}
\end{table}

\paragraph{Definition}
Much attention has been given to query rewriting techniques, first introduced by \cite{ma-etal-2023-query}. The idea is that when the user query is tested against the knowledge base for a similarity search, 
the comparison is often performed among fairly different texts: the query is usually a short and dense question, while chunks in the KB can be long and complex. Query rewriting techniques aim to reduce this delta by converting the query into a text with a structure more similar to the target chunks.  Hyde \cite{gao-etal-2023-precise}, consisting in substituting the query with a short paragraph that answers it, has emerged as one of the best performing techniques \cite{wang-etal-2024-searching}.

\paragraph{Enhanced Implementation}
We forced the Enhanced system to perform Hyde query rewriting. Each user query is automatically rewritten before retrieval  by prompting \texttt{gpt-4o}.

\paragraph{Agentic Implementation}
We design a prompt to make the agent aware that rewriting might  (Appendix \ref{app:prompt}). Once the Agent has chosen to use the RAG tool, it can decide to perform this step.

\paragraph{Experimental Settings}
We run all the queries in the test sets of the four datasets against the two systems. 
In those cases where the Agentic setting did not perform query rewriting, we calculate the retrieval metric on the original query. 

\paragraph{Evaluation metric}
All queries in the four datasets come with annotations on the ground truth documents they should be linked to.
When evaluating the quality of the retrieved chunks, we use the Nomarlized Discounted Cumulative Gain NDCG@10 \citep{10.1145/582415.582418}. More details are reported in Appendix~\ref{app:ndcg}.

\paragraph{Results}

Results in Table~\ref{tab:rewriting} show that the Agentic setting performs better than Enhanced, with an average gain of $+2.8$ NDCG@10 points. We attribute this to the flexibility of the former, which can dynamically decide whether to perform rewriting, and how. Results suggest that query rewriting is beneficial in general, and that an adaptive approach that decides what to do on a case-by-case basis is the most effective one. \textcolor{black}{Rewriting is maximally useful in general when the user query is very different from a question format (FEVER). It can be observed that when user queries can be of any kind (NQ), the flexibility of the Agent allows it to outperform Enhanced settings ($+7.8$ points). On the other hand, for specific-domain settings, they perform equally (FIQA). Interestingly, for both IR/E tasks the Agent outperforms Enhanced settings by the same delta ($+2$ and $+1.5$ points), suggesting that when RAG is used for information extraction the flexible rewriting is desirable.}
Examples are reported in Appedix~\ref{app:example_rewrite}.




\subsection{Document List Refinement}

\paragraph{Definition}
Previous work has shown how the retrieval step may include partially irrelevant or noisy results, and proposed approaches to improve the selection process via reranking strategies \cite{sachan-etal-2022-improving, sun-etal-2023-chatgpt, qin-etal-2024-large}. 
Reranking consists in sorting the retrieved chunks, selecting a subset of highly relevant ones with respect to the user query. Other methods such as CRAG \citep{DBLP:journals/corr/abs-2401-15884} have been proposed, highlighting the importance of such step.

\paragraph{Enhanced Implementation}
We experiment with this dimension by using an ELECTRA-based reranker\footnote{\url{https://huggingface.co/naver/trecdl22-crossencoder-electra}} \cite{déjean2024thoroughcomparisoncrossencodersllms} on the list of the $20$ most similar documents for each user query.

\paragraph{Agentic Implementation}
The Agentic RAG system can inherently attempt to consider a more suitable context when needed. Specifically, the agent may trigger additional retrieval rounds and adapt the query formulation as it deems appropriate, allowing it to iteratively obtain more relevant context.
We calculate the metric on the last reformulation of the query that the Agent uses for the RAG tool, which directly precedes answer generation.

\paragraph{Experimental Settings}
We run all queries in the FIQA and CDQStack-En test sets against the two systems to assess performance on both QA and IR/E tasks. We did not consider NQ and FEVER due to their size. As detailed in Section~\ref{sec:cost}, Agentic RAG would take >7 days on each of their KBs.

\paragraph{Evaluation metric}
As in query rewriting, we utilize NDCG@10, leveraging the ground truth links between queries and documents.

\paragraph{Results}
In the Enhanced setting, re-ranking has a substantial positive impact on performance. In contrast, the Agentic RAG setting gains no benefit from iterating the retrieval step.
On average, in those cases in which the agent decided to perform again the retrieval (10\% of the times), 53\% of the retreived documents remain the same (example reported in Appendix~\ref{app:example_rerank}). \textcolor{black}{This highlights that once the model has taken a decision, it is not likely to reconsider it. Even when that happens, the process leads to a second retrieval step that only in one case out of two modifies the retrieved documents with respect to the previous step.}





\begin{table}[]
\centering
\renewcommand{\arraystretch}{1.2}
\small
\begin{tabular}{|l|c|c|c|} \hline
& \textbf{QA} & \textbf{IR/E} & \\ \hline
    \textbf{Settings}     & \textbf{FIQA} & \textbf{CQA-EN} & \textbf{AVG}   \\
\hline
naïve & 45.0  & 46.0           & 45.5  \\
\hline
enhanced w/o rewriting & 49.0   & 47.0          & \underline{48.0}  \\
\hline
enhanced  with rewriting & 51.0   & 48.0           & \textbf{49.5}  \\
\hline
agent    & 43.4   & 44.4         & 43.9  \\
\hline
\end{tabular}
\caption{\textbf{Document list refinement performances} in terms of NDCG@10. 
Since Agentic retrieval might indirectly performs query rewriting, we report results with and without rewriting for the Enhanced setting.}
\label{tab:reranking}
\end{table}

\begin{figure}[t]
  \includegraphics[width=\linewidth]{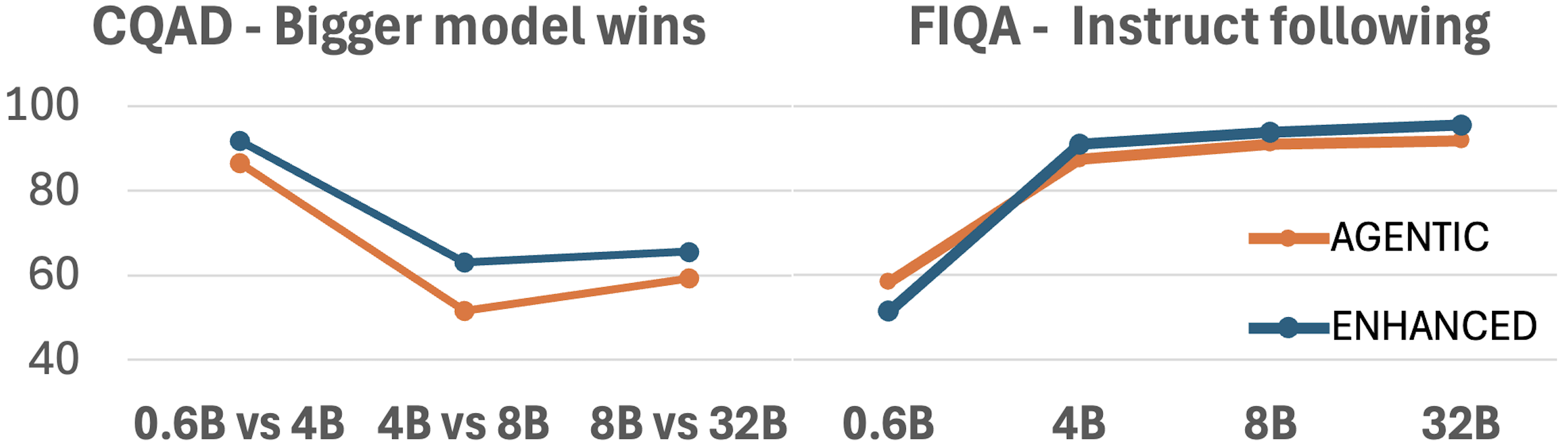}
  \caption {\textbf{Changing underlying LLM performances (Qwen)}. For CQADupStack-EN (\textbf{left}), the metric is based on a pairwise analysis, calculated as the \% ratio of times the larger model's answer is better than the smaller counterpart. For FIQA (\textbf{right}), the metric is the overall \% ratio of the classification metric (1 if the answer follows the instruction, 0 otherwise).   Both metrics are calculated via LLM-as-a-Judge (Selene-70B).}
\label{fig:selene}
\end{figure}

\subsection{Underlying LLM} \label{sec:underlying_llm}

\paragraph{Definition}
Both Agentic and Enhanced settings are highly impacted by the choice of the underlying LLM, as different models produce different answers even when provided with the same query and retrieved context. Furthermore, the role of the generator is particularly critical in Agentic RAG, where the model must not only produce the final answer but also make decisions at each stage of the workflow.
We are interested in quantifying the impact that a weaker generator has on the system, compared to what a stronger one would have. 

\paragraph{Enhanced and Agentic Implementation}
To assess this effect, we tested four generators of varying capability, namely \texttt{Qwen3-0.6B} (without thinking), \texttt{Qwen3-4B}, \texttt{Qwen3-8B}, and \texttt{Qwen3-32B} \citep{yang2025qwen3technicalreport}.
We define the generator capability based on the self-reported performances (Table~\ref{tab:generator} in Appendix). We do not aim to assess which model performs best as a generator, but how the overall RAG performances are impacted by generators of different power.
We utilise Enhanced RAG with rewriting and reranking, and the Agentic settings with the system prompt defined in Appendix~\ref{app:prompt}.

\paragraph{Experimental Settings}
We run all queries in the test sets of FIQA (QA) and CDQStack-En (IR/E) against the two systems. We did not consider NQ and FEVER due to their KB size.
For both systems, we set the retrieval chunks number  to $5$.

\paragraph{Evaluation metric}
We evaluate the quality of the final answers generated by both systems using automatic metrics, employing the LLM-as-a-judge paradigm. Out of the many approaches proposed in the literature \citep{kim2024prometheusinducingfinegrainedevaluation, wang-etal-2025-direct-judgement, es-etal-2024-ragas}  we select Selene-70B \citep{alexandru2025atlaseleneminigeneral}, a fine-tuned version of Llama-3.3-70B-Instruct, as its smaller version scores among the top ones in the LLM-as-a-Judge \citep{judge-arena-ai}.
Furthermore, we analyzed its reliability in evaluating our two selected datasets. First, Selene judgments has been shown to closely match human evaluations on financial QA tasks by its authors, which is relevant for the FIQA dataset. Therefore, we could adopt the binomial 0–1 classification metric for this dataset, shown to be most human-aligned option.
On the other hand, for CQADupStack-En, such alignment has not been demonstrated. Therefore, we performed manual annotation on a subset of the testing data ($5\%$, $312$ answer pairs in total), and calculated the agreement rate between the two human annotators and the automatic metric. We select the pairwise metric (given two answers of two models, select the best one). Inter-annotator agreement rate (ratio of times they both chose A or B) is $71.9\%$, while the human-model agreement is $65.4\%$. Manual annotation took an average of $1.5$ minutes per pair, resulting in $15.5$ hours overall. The annotation guidelines are reported in the Appendix.
To summarize the impact of underlying model changes, we calculate the ratio of times the larger model wins over the smaller counterpart.

\paragraph{Results}
We aim to assess whether Enhanced and Agentic systems exhibit distinct performance patterns across model scales.
Figure~\ref{fig:selene} reports the resulting average scores for both datasets, which show that the two systems do not present significant differences in patterns when changing the underlying LLM. In fact, the performance increase in FIQA follows the same distribution both Enhanced and Agentic RAG, and the same is true for the ratio of times in which the bigger model is preferred over the smaller one in CQADupStack-En settings. Full results are reported in Appendix~\ref{app:selene}.

\section{Cost and Time} \label{sec:cost}




\paragraph{Fixed costs for Enhanced and Agentic settings}
The two systems share some fixed costs due to hardware requirements. We utilize for both settings a \texttt{t3.large} ec2 AWS instance (0.09 \$/h per on-demand usage) to instantiate a relational database with vector search capabilities (\texttt{pgvector}\footnote{\url{https://github.com/pgvector/pgvector}}).  We implement the RAG application backend on a \texttt{t2.medium} (0.05 \$/h).
We test open LLMs in a proprietary 8$\times$a40 cluster (46GB). We run Qwen3 0.6B, 4B, and 8B on a single gpu, while the 32B version on 4$\times$ a40. A similar setting on AWS can be \texttt{g4ad.8xlarge}, which costs 1.9 \$/h. The time and cost related to retrieval are the same in both settings. We use OpenAI \texttt{text-embedding-3-small} as default embedder model with cosine similarity.
Enhanced RAG does re-ranking with a 300M-parameter model on the same cluster as the LLMs, with a negligible added cost.

\paragraph{Runtime costs and number of tokens}
We approximate the runtime cost of RAG systems by the number of processed input and output tokens. This hardware-agnostic metric enables cost estimation across different deployment environments given on model throughput and hourly pricing.
We analyse token usage for both GPT and Qwen models to quantify cost differences between Enhanced and Agentic settings. We also report end-to-end latency—the time from receiving a query to returning the final answer. 
Table~\ref{tab:costs} (Appendix) reports a summary of time and tokens per approach.\\
Overall, we find that Agentic settings are more expensive and more time-consuming than Ehanced settings, requiring an average of $3.3\times$ more input tokens and $1.9\times$ more output tokens among datasets, as well as $1.5\times$ more time.\\
For valid queries in Enhanced RAG, we find that roughly 45–50\% of the total time is spent generating the answer, a similar proportion is spent on query rewriting, 0–5\% on retrieval, and 0–2\% on document re-ranking. The dominant factor in latency is the LLM calls. Therefore, any performance optimization should focus primarily on that.

\section{Conclusion}

Our experimental comparison reveals that neither Enhanced nor Agentic RAG is universally superior.
First, we observe that in well-defined domains with highly structured user behavior, Agentic RAG excels at \textbf{handling user intent,} thanks to its ability to understand the user query. However, in broader or noisier domains, our Enhanced RAG routing system proves more reliable. \
textcolor{blue}{Developers should consider using the Agentic approach when possible, considering it does not require any manually crafted example to run.}
Second, with respect to \textbf{alignment of the query to} the structure and semantics of the \textbf{documents} in the KB, Agentic RAG outperforms Enhanced RAG retrieval quality. Its dynamic use of query rewriting allows for retrieval of more relevant context.  
Third, we found that when Agentic RAG selects certain documents, it is not as good as the \textbf{re-ranking} done by Enhanced RAG at selecting just the most meaningful docs. 
Fourth, we observe that changing the \textbf{underlying LLM} produce the same changes in performance in both settings:
as the LLM becomes larger, performance improves at comparable rates.
Our \textbf{cost} analysis highlights that Agentic RAG is systematically more expensive—up to 3.6 times more—due to additional reasoning steps and repeated tool calls. This cost difference should not be overlooked: a well-optimized Enhanced RAG can match or exceed Agentic performance while remaining more efficient. \\
\textcolor{black}{In summary, developers should consider combining the two approaches to achieve the best performances, taking into accout the increase in cost. The Agentic approach suits best user-intent routing (even without any manually crafted examples) and query rewriting. On the other hand, our results suggest that integrating an explicit re-ranking step into Agentic pipelines could provide substantial gains.}

\bibliography{custom}

\appendix

\section{Appendix}

\subsection{NDCG metric} \label{app:ndcg}

When evaluating the quality of the retrieved chunks, we use the Nomarlized Discounted Cumulative Gain NDCG@10 \citep{10.1145/582415.582418}. 
NDCG stands for Nomarlized Discounted Cumulative Gain \citep{10.1145/582415.582418} and is one of the most common metrics used to assess the effectiveness of a ranking model.
NDCG at cutoff $K$ is defined as:

\begin{equation}\label{eq:ndcg@k}
\text{NDCG@K} \equiv \frac{\text{DCG@K}}{\text{maxDCG@K}}
\end{equation}\label{eq:ndcg}\\
where $\text{maxDCG@K}$ is the maximum DCG@K that can be obtained from the given relevance labels, and where DCG@K is defined as:

\begin{equation}\label{eq:ndc@k}
\text{DCG@K} \equiv \sum_{i=1}^{K} \frac{2^{l_i} - 1}{\log(1 + i)}
\end{equation}\label{eq:ndc}\\
where $l_i$ is the relevance label (the ground truth label) of the document in position $i$ in the rank. Since Equation~\ref{eq:ndcg} is always positive, Equation~\ref{eq:ndc} is a number bounded between 0 and 1, where NDCG@K equal to 1 means that we have a perfect ranked list.

\subsection{Invalid query generation} \label{app:invlaid_gen}
For each dataset, the invalid queries are generated by prompting gpt-4o with the following prompt:

\newtcolorbox{promptbox}{
  colback=gray!8,
  colframe=gray!60,
  boxrule=0.48pt,
  arc=1mm,
  left=2mm, right=2mm, top=1mm, bottom=1mm,
  breakable,
  enhanced,
  fontupper=\ttfamily\small,
}

\begin{promptbox}
You are a data augmentation assistant focused on GENERATING OFF-DOMAIN QUERIES.\\\\
You are given VALID EXAMPLES representing the domain of a knowledge base (KB). Produce EXACTLY {n} distinct
user queries that are OFF-TOPIC relative to the examples, thus NOT answerable by retrieval over that KB.
\\\\
REQUIREMENTS:\\
  - Each query 5-18 words, self-contained, concrete.\\
  - Benign topics only; no sensitive or disallowed content.\\
  - Must NOT overlap semantically with examples or with each other.\\
  - Avoid meta references (no mentions of 'examples', 'dataset', 'KB', 'instructions').\\
  - Diverse domains (finance, cooking, travel, biology, art, sports, etc.).
\\\\
OUTPUT STRICT JSON ONLY (no extra text):
\{\{"invalid\_queries": ["query 1", "query 2", ... (total 250) ]\}\}
\\\\
VALID EXAMPLES:\\\{examples\_formatted\}\\
Return ONLY the JSON now.
\end{promptbox}

To quantify the distances between the two group of queries ("VALID" and "INVALID") for each dataset, we calculate the cosine similarity between embeddings calculated using \texttt{Qwen/Qwen3-Embedding-0.6B}:
\begin{itemize}
    \item between each \textit{valid} query and the other $249$ \textit{valid} queries (\textit{V-V}),
    \item between each \textit{valid} query and the other $250$ \textit{invalid} queries (\textit{V-I}).
\end{itemize}
Higher deltas between \textit{V-V} and \textit{V-I} values mean that queries are quite distant, as desired.
We average the similarities \textit{V-V} and \textit{V-I} and found good discrimination on CQAD-EN (\textit{V-V} 0.449, \textit{V-I} 0.276), NQ (\textit{V-V} 0.339, \textit{V-I} 0.266), FIQA NQ (\textit{V-V} 0.404, \textit{V-I} 0.257). This distance is much lower for FEVER, which indeed we excluded (\textit{V-V} 0.289, \textit{V-I} 0.225).

\subsection{Prompts} \label{app:prompt}
\subsubsection{Agentic RAG}
The Agentic RAG is defined by three nodes: the orchestrator, the answer node, and the RAG node. Here we report the orchestrator prompt:

\begin{promptbox}
 Context \\
You are a powerful Al agent. Your task is to \{task\_description\}\\ \\
 Inputs
\\•query: \{query\}
\\•previous\_tool\_results: \{previous\_tool\_results\}\\ \\
 Guidelines
To do complete your task, you can take several actions:\\
• answer: \{answer\_node\_description\}
\\• document\_retrieval:\ \\\{retrieval\_node\_description\}\\ \\
 Important
\\• Don't repeat actions unnecessarily.
\\• When composing the query for RAG search you must always write a passage to answer the input query provided by the user.
\\• If you think that the retrieved documents in the previous\_tool\_results can be further improved, you can rewrite the query and call document\_retrieval again.
\\• If you see the relevant retrieved documents already in the previous\_tool\_results, you should proceed with answering!\\ \\
 Metadata:
\\• current reasoning step: \{current\_reasoning\_step\}
\\• maximum reasoning steps: \{max\_reasoning\_steps\}
\end{promptbox}

For each evaluation dataset, the prompt if filled with the following:

\begin{promptbox}
FIQA\\
task\_description:\\
help users on financial issues using the tools at your disposal.\\ \\
answer\_node\_description:\\
provide the final answer to the user, utilizing the retrieved documents if necessary. Choose this action when you have gathered enough information from the previous\_tool\_results.\\ \\
retrieval\_node\_description:\\
retrieve documents from a knowledge base of financial information.
Choose this to ground answers on external sources
\end{promptbox}

\begin{promptbox}
NQ\\
task\_description:\\
help users using the tools at your disposal.\\ \\
answer\_node\_description:\\
provide the final answer to the user, utilizing the retrieved documents if necessary. Choose this action when you have gathered enough information from the previous\_tool\_results.\\ \\
retrieval\_node\_description:\\
retrieve documents from a knowledge base of potentially useful information. Choose this to ground answers on external sources.
\end{promptbox}

\begin{promptbox}
FEVER\\
task\_description:\\
help users verifying their queries on factual knowledge. For each user claim, you determine if it is supported or refuted by the world-knowledge on the topic.\\ \\
answer\_node\_description:\\
provide the final answer to the user, utilizing the retrieved documents if necessary. Choose this action when you have gathered enough information from the previous\_tool\_results.\\ \\
retrieval\_node\_description:\\
retrieve documents from a knowledge base of factual information. Choose this to ground answers on gold-standard knowledge. Be careful, your knowledge might be outdated.
\end{promptbox}

\begin{promptbox}
CQADUPSTACK-EN\\
task\_description:\\
help users finding blog posts about English grammar issues that are related to the user query. You have to find the relevant documents, summarize them, and inform the user about them.\\ \\
answer\_node\_description:\\
provide a summary of the retrieved blog posts to the user. Choose this action when you have gathered enough information on grammar blog posts from the previous\_tool\_results.\\ \\
retrieval\_node\_description:\\
retrieve blog posts from a knowledge base of English grammar issues. Choose this to find relevant blog posts. Never use it for queries not related to English grammar issues.
\end{promptbox}

The answer node is defined as follows:

\begin{promptbox}
FIOA \& NQ\\
Context:\\
You are a question-answering assistant. You are provided the original user query, a list of tools called and their results, a draft answer, and the conversation history. Provide the final answer to the user.\\ \\
Inputs:\\
question: \{query\}\\
previous\_tool\_results: \{previous\_tool\_results\}\\
draft\_answer: \{draft\_answer\}\\ \\
Guidelines:\\
- be consistent with previous responses in the conversation.\\
- if a follow-up question references a previous conversation, use the conversation history to provide context.
\end{promptbox}

\begin{promptbox}
FEVER\\
 Context:\\
You are a question-answering assistant. You are provided the original user query, a list of tools called and their results, a draft answer, and the conversation history. Provide a summary of the grounding reasons and evidence.\\ \\
Inputs:\\
question: \{query\}\\
previous\_tool\_results: \{previous\_tool\_results\}\\
draft\_answer: \{draft\_answer\}\\ \\
Guidelines:\\
- be consistent with previous responses in the conversation.\\
- if a follow-up question references a previous conversation, use the conversation history to provide context.\\
- summarize the reasons that support or contradict the statement.
\end{promptbox}

\begin{promptbox}
CQADUPSTACK-EN\\
 Context:\\
You are a blog posts summarizer. You are provided the original user query, a list of tools called and their results, a draft answer, and the conversation history. Provide a bullet point list of the relevant blog posts related to the user query, with one bullet point per blog post. Each bullet point must include: i) a short title (max 10 words); ii) a one-sentence summary (max 15 words) of the blog post; iii) the reference to the document. You do not answer the question. Instead, you provide an overview of the retrieved blog posts to the user. Your objective is to make the user aware of the relevant blog posts related to their query.\\ \\
 Inputs:\\
question: \{query\}\\
previous\_tool\_results: \{previous\_tool\_results\}\\
draft\_answer: \{draft\_answer\}\\ \\
Guidelines:\\
- be consistent with previous responses in the conversation.\\
- introduce the bullet point list with a short sentence such as ``Here are some blog posts that might be relevant to your query:''.\\
- do not answer the question directly. Your goal is to summarize the retrieved blog posts.
\end{promptbox}

The RAG node does not have a specific system prompt.\\
\paragraph{Query rewriting}Query rewriting is performed with the following prompt: 
\textit{"Convert the user query into a \{type\_of\_doc\}"}, where \textit{type\_of\_doc} differs based on the dataset (\textit{"longer blog post"} for CQADupStack, \textit{"passage to answer it"} for FiQA and NQ, \textit{"longer factual statement"} for FEVER).

\subsubsection{Enhanced RAG}
For each of the four evaluation dataset a different system prompt is defined: 

\begin{promptbox}
FIQA\\
You are an advanced AI assistant expert in financial topics. You are provided with a question from a user and some potentially useful context. You are able to use the useful context to further inform your responses. This context is divided into chunks drawn from different documents. Remember that the useful context is your cheatsheet, so you must always avoid mentioning it to the user. Never mention the context.\\ \\
**Very Important**:\\
- You MUST NEVER invent documents that are not mentioned in the 'useful context'.\\
- if you do not see a chunk matching the user's question, do not mention it.\\
- avoid hallucinating data. Answer truthfully and forthrightly, stating only what you can derive from the provided context or from general knowledge.
\end{promptbox}

\begin{promptbox}
NQ\\
You are an advanced AI assistant. You are provided with a query from a user and some potentially useful context. You are able to use the useful context to further inform your responses. This context is divided into chunks drawn from different documents. Remember that the useful context is your cheatsheet, so you must always avoid mentioning it to the user. Never mention the context.\\ \\
**Very Important**:\\
- you must never invent documents that are not mentioned in the useful context.\\
- if you do not see a chunk matching the user's question, do not mention it.\\
- avoid hallucinating data. Answer truthfully and forthrightly, stating only what you can derive from the provided context or from general knowledge.
\end{promptbox}

\begin{promptbox}
FEVER\\
You are an advanced AI assistant. Your task is to help the user verify their queries on factual knowledge. For each user claim, you determine whether it is supported or refuted by the world-knowledge on the topic. You are provided with a query from the user and some potentially useful context. You are able to use the useful context to further inform your responses. This context is divided into chunks drawn from different documents. Remember that the useful context is your cheatsheet, so you must always avoid mentioning it to the user. Never mention the context.\\ \\
**Very Important**:\\
- you must never invent documents that are not mentioned in the useful context.\\
- if you do not see a chunk matching the user's question, do not mention it.\\
- avoid hallucinating data. Answer truthfully and forthrightly, stating only what you can derive from the provided context or from general knowledge.
\end{promptbox}

\begin{promptbox}
CQADUPSTACK-EN\\
You are a blog posts summarizer. Your task is to help users find blog posts about English grammar issues that are related to the user query. You are provided with relevant documents; summarize them and inform the user about them. Provide a bullet point list of the relevant blog posts related to the user query, with one bullet point per blog post. Each bullet point must include: i) a short title (max 10 words); ii) a one-sentence summary (max 15 words) of the blog post; iii) the reference to the document. You do not answer the question. Instead, you provide an overview of the retrieved blog posts to the user. Your objective is to make the user aware of the relevant blog posts related to their query.\\ \\
**Very Important**:\\
-be consistent with previous responses in the conversation.\\
- introduce the bullet point list with a short sentence such as ``Here are some blog posts that might be relevant to your query:''.\\
- do not answer the question directly. Your goal is to summarize the retrieved blog posts.
\end{promptbox}

\paragraph{Query rewriting} Query rewriting is performed with the following prompt: 
\textit{Please write a passage to answer the question. \textbackslash n Question: \{user\_query\}\textbackslash n Passage:"}.

\subsection{Routing system details} \label{app:router}
\begin{figure}[t]
  \includegraphics[width=\linewidth]{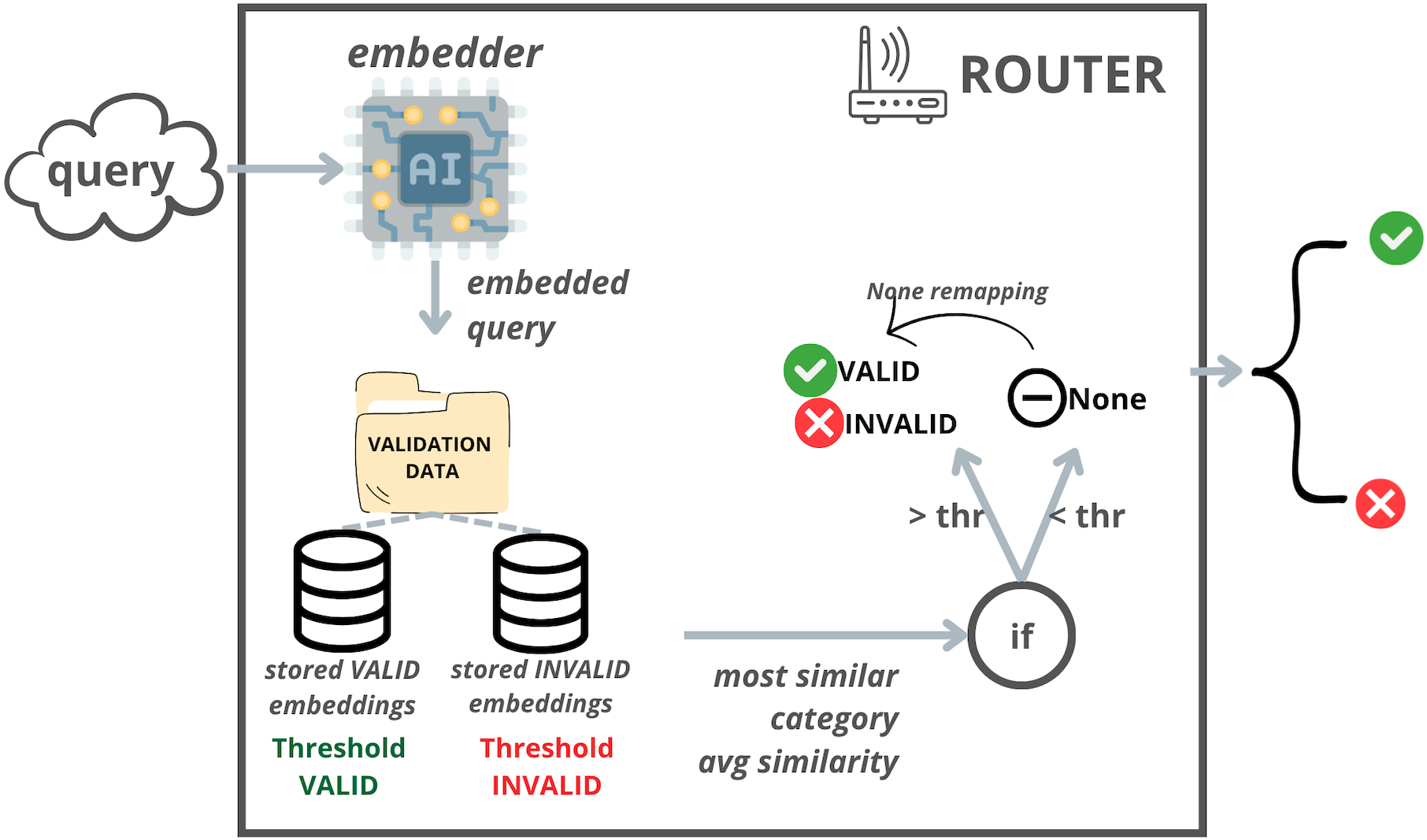}
  \caption{\textbf{Schema of the routing system utilized for Enhanced}.
Two query classes are defined—valid and invalid—each represented by embedded examples from a validation set. For each incoming query, the router embeds it, retrieves the top-20 most similar examples via cosine similarity, and selects the class with the highest average similarity. If the mean similarity of the selected class exceeds a predefined threshold, that label is assigned; otherwise, the router returns \textit{None}, which can be further remapped according to business logic.
} \label{fig:router}
\end{figure}

The schema of the routing system we implement for Enhanced RAG settings is described in Figure~\ref{fig:router}.
The routing mechanism relies on example-based classification. Queries are compared to two reference sets: valid and invalid. If a query is sufficiently similar to the valid set, it is processed; otherwise, it is rejected. The key challenge is determining what “sufficiently similar” means, i.e., selecting an appropriate similarity threshold.

\paragraph{Threshold selection} 

Threshold selection can be approached in multiple ways (e.g., random search, linear models, classification algorithms).
If class definitions are clear and well-separated, threshold tuning becomes less critical, since queries will naturally cluster around their correct class. In practice, however, class boundaries often contain noise, making the threshold an essential safeguard against misclassification.

\subsection{Underlying LLM evaluation metrics} \label{app:selene}

The results for the metrics calculated by means of Selene-70B are reported in Table~\ref{tab:all_selene_fiqa} (FIQA) and Figure~\ref{fig:all_selene_cqa} (CQADupStack-EN). Here we report the evaluation guidelines for CQADupStack-English used by human annotators:

\begin{promptbox}
GENERAL DESCRIPTION\\
You are evaluating a system that, given a user query, retrieves a list of similar queries previously submitted by other users. These past queries are called documents. The system output must be a list of bullet points, each representing one and only one document. An ideal bullet point contains:\\
- a title of at most five words,
\\- a brief description of the retrieved document,\\
- a reference to the document number.\\ \\
Important:
the documents do not contain answers to the user's query; they are actual queries submitted in the past. If the system response includes a document description that explains the grammar question rather than restating the query itself, this is incorrect.\\ \\
ANNOTATION:\\
For each query, there are two system responses, A and B. Your task is to select the better response based on the criteria above. If you are unsure, you may write ``I don't know.'' If both responses are of equal quality, write ``same.''\\ \\
Common situations include:\\
- the document description contains an answer rather than the original query (serious error).\\
- one of the listed documents is irrelevant to the query (serious error).\\
- if the two responses differ in the number of documents and all documents are relevant, always choose the one with more documents.\\
- the response mixes the document number with the title (error).\\
- the document description is excessively long (error).\\
- the document number reference is missing (error).
\end{promptbox}

Prompt used for Selene 70B for CQADupStack-English, based on the one released with the original paper \cite{alexandru2025atlaseleneminigeneral}:

\begin{promptbox}
EVALUATION TASK\\
You are a helpful assistant evaluating the quality of responses for a given instruction. Your goal is to select the best response for the given instruction. Two responses are provided, Response~A and Response~B, generated by two different AI chatbots. You must select either Response~A or Response~B. Do not state that both or neither are good.\\ \\
Here are some rules of the evaluation:\\
(1) You should prioritize evaluating whether the response satisfies the provided rubric. Then consider its helpfulness, accuracy, level of detail, harmlessness, etc.\\
    (2) Responses should NOT contain more/less than what the instruction asks for, as such responses do NOT precisely execute the instruction.\\
    (3) You should avoid any potential bias and your judgment should be as objective as possible. Here are some potential sources of bias:\\
    - The order in which the responses were presented should NOT affect your judgment, as Response A and Response B are **equally likely** to be the better.\\
    - The length of the responses should NOT affect your judgement, as a longer response does not necessarily correspond to a better response. When making your decision, evaluate if the response length is appropriate for the given instruction.
\\ \\
 Your reply should strictly follow this format:\\ \\
    **Reasoning:**  <feedback evaluating the responses>\\
    **Result:**  <A or B>\\\\
    Here is the data.\\\\
    Instruction:
    \{full\_prompt\_for\_answer\_gene\}\\
    Response A:
    \{assistant\_response\_a\}\\
    Response B:
    \{assistant\_response\_b\}\\
    Score Rubrics:
    [Which of the following responses better handles the Instruction - A or B?]
\end{promptbox}

Prompt used for Selene 70B for FIQA:
\begin{promptbox}
    You are tasked with evaluating a response based on a given user input and binary scoring rubric that serves as the evaluation standard. Provide comprehensive feedback on the response quality strictly adhering to the scoring rubric, followed by a binary Yes/No judgment. Avoid generating any additional opening, closing, or explanations.\\\\
    Here are some rules of the evaluation:\\
    (1) You should prioritize evaluating whether the response satisfies the provided rubric. The basis of your score should depend exactly on the rubric. However, the response does not need to explicitly address points raised in the rubric. Rather, evaluate the response based on the criteria outlined in the rubric.\\\\
    Your reply must strictly follow this format:\\\\
    **Reasoning:** <Your feedback>

    **Result:** <Yes or No>

    Here is the data:

    Instruction:
    \{full\_prompt\_for\_answer\_gen\}\\

    Response:\\
    \{assistant\_response\} \\ \\
    Score Rubrics:\\
    Does the response correctly and completely answers the question posed in the instruction? Answer Yes or No.\\
    Yes: The response accurately and completely addresses the question posed in the instruction \\
    No: The response fails to accurately and completely address the question posed in the instruction.
\end{promptbox}

\begin{table}[]
\centering
\small
\renewcommand{\arraystretch}{1.2}
\begin{tabular}{|c|rrr|r|}
\hline
            \textbf{Model}          & \textbf{General}      & \textbf{Align} & \textbf{Reason} &   \textbf{AVG}     \\
           Qwen3           & GPQA-D & IFEVAL     & AIME '24  & \\
\hline
0.6B    & 22.9         & 54.5       & 3.4       & 26.9 \\
\hline
4B      & 55.9         & 81.9       & 73.0        & 70.3 \\
\hline
8B       & 62.0           & 85.0         & 76.0        & 74.3 \\
\hline
32B       & 68.4           & 85         & 81.4        & 78.3 \\ \hline
\end{tabular}
\caption{Performances of the models selected to analyse the impact of the underlying LLM as reported by \citet{yang2025qwen3technicalreport}. The selected benchmarks are Graduate-Level Google-Proof QA Diamond \citep{rein2023gpqagraduatelevelgoogleproofqa}, Instruction Following Eval \citep{zhou2023instructionfollowingevaluationlargelanguage}, and the American Invitational Mathematics Examination \citep{AIME2024}.}
\label{tab:generator}
\end{table}
\normalsize

\begin{figure}[t]
\centering

\begin{minipage}[t]{0.48\textwidth}
  \centering
  \small
  \renewcommand{\arraystretch}{1.2}
  \begin{tabular}{|c|l|l|}
    \hline & \multicolumn{2}{c|}{\textbf{FIQA}} \\
    \hline
    \textbf{Qwen} & enhanced & agentic \\
    \hline
    \textbf{0.6B}  & 51,4 & 58,4 \\
    \hline
    \textbf{4B}   & 90,9 & 87,3 \\
    \hline
    \textbf{8B}   & 93,7 & 91 \\
    \hline
    \textbf{32B}  & 95,5 & 91,8 \\
    \hline
  \end{tabular}
   \captionof{table}{Classification metric based on Selene-70B for \textbf{FIQA}.}
  \label{tab:all_selene_fiqa}
\end{minipage}
\vfill
\begin{minipage}[t]{0.48\textwidth}
  \centering
  \includegraphics[width=\linewidth]{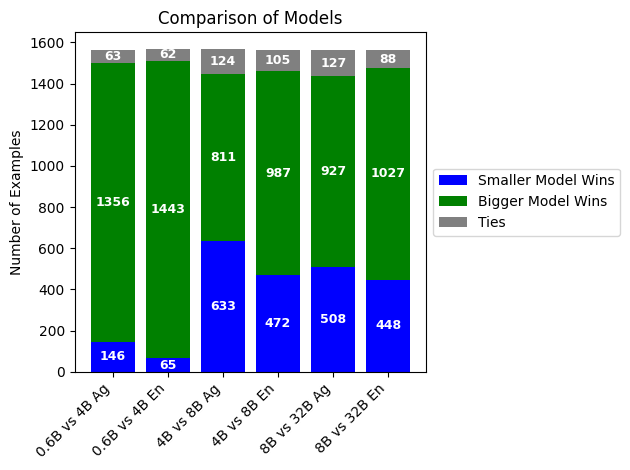}
  \caption{Pairwise metric based on Selene-70B for \textbf{CQADupStack-EN}.}
  \label{fig:all_selene_cqa}
\end{minipage}
\end{figure}





\begin{table*}[]
\centering
\small
\renewcommand{\arraystretch}{1.1}
\begin{tabular}{|l|p{6cm}|p{6.8cm}|}
\hline
\textbf{Framework} & \textbf{pros} & \textbf{cons} \\
\hline

smolagents &
minimal; codeagent & lack of low-level abstraction; lower control and security issues because of \texttt{codeagent} (can be handled) \\
\hline
langgraph & graph abstraction; many integrations; many resources for implementing popular patterns & \\
\hline
llamaindex & many integrations; many resources for implementing popular patterns & high-complexity codebase \\
\hline
pocketflow & minimal; graph abstraction & need to perform implementation \\
\hline
crewAI &  & focus on multi-agents \\
\hline
autogen & mature ecosystem & high complexity \\
\hline
pydanticAI & type safety; Python oriented & early stages of development; requires learning some concepts \\
\hline
atomic agents & modular & early stages of development, limited documentation \\
\hline

\end{tabular}
\caption{Comparison of frameworks considered for Agentic RAG implementation. 
The reported “pros” and “cons” are defined within the scope of this work—namely, the construction of an agent equipped with a single RAG tool—and should not be considered exhaustive or universally applicable.
We considered SmolAgents \citep{smolagents}, LangGraph \citep{langgraph_repo}, LlamaIndex \citep{Liu_LlamaIndex_2022}, PocketFlow \citep{pockeyflow_repo}, CrewA \citep{crewai}, AutoGen \citep{autogen_repo}, PydanticAI \citep{pydantic-ai}, and Atomic Agents \citep{atomic_agents_repo}, offering different advantages and reflecting different design choices.
For this work, we select PocketFlow, a lightweight framework that offers a simple graph-based abstraction.
}
\label{tab:agent_imp_frameworks}
\end{table*}


\begin{figure*}[t]
  \includegraphics[width=0.48\linewidth]{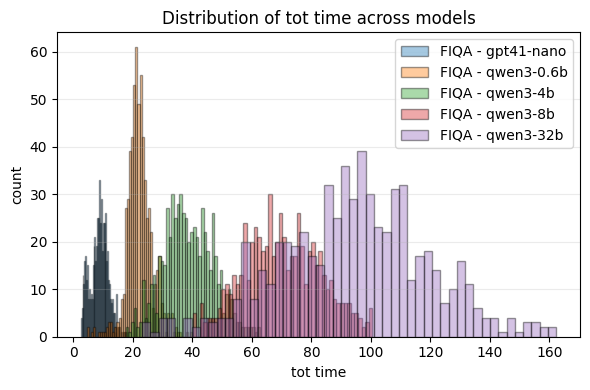} \hfill
  \includegraphics[width=0.48\linewidth]{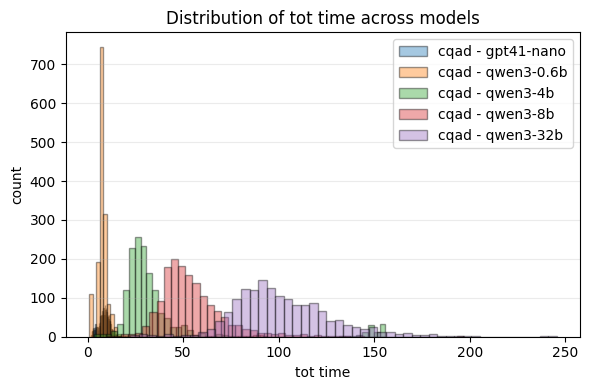}\hfill
  \includegraphics[width=0.48\linewidth]{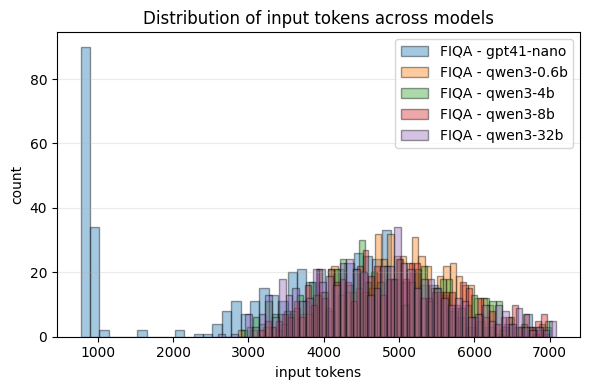} \hfill
  \includegraphics[width=0.48\linewidth]{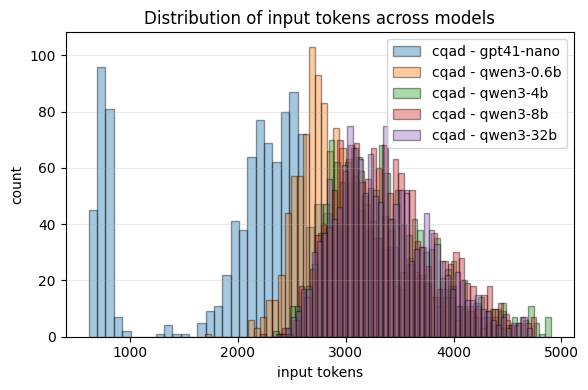}\hfill
  \includegraphics[width=0.48\linewidth]{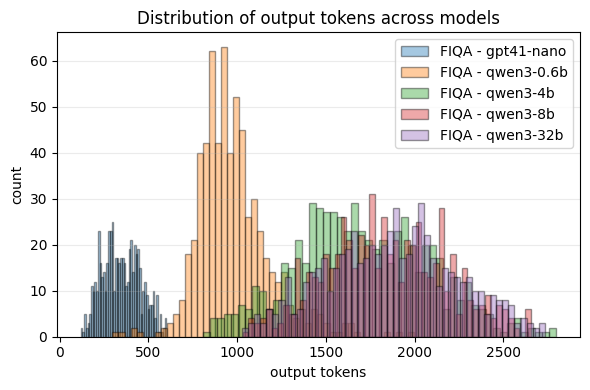} \hfill
  \includegraphics[width=0.48\linewidth]{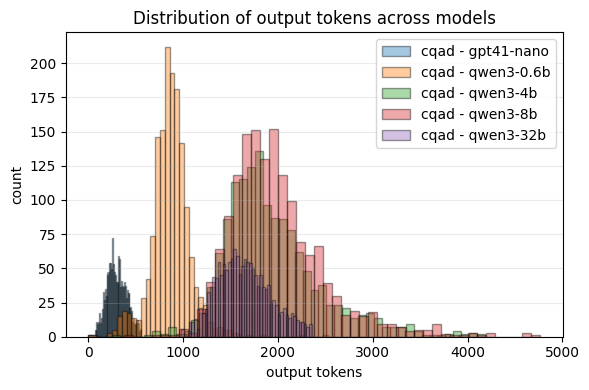}
  \caption {Overall computational cost and token usage for each model in the \textbf{Agentic} setting when processing a single user query. Qwen3 4B, 8B, and 32B operate in thinking mode, whereas the 0.6B variant is run without it. Qwen3 0.6B, 4B, and 8B are executed on a single NVIDIA A40, while Qwen3 32B is run on 4$\times$A40 GPUs. \\
\textbf{Top:} Distribution of total latency, measured from the moment the system receives the query to the moment the final answer is produced. Within the Qwen family, Qwen3-0.6B achieves the lowest latency due to its smaller size and the absence of thinking mode.
\textbf{Middle:} Average number of input tokens. This value increases slightly with model size. 
The peak of low input-token values for GPT-4.1-nano arises because the model frequently opts not to use the RAG tool, thereby reducing the number of required reasoning steps.\\
\textbf{Bottom:} Average number of output tokens. Here, the largest difference emerges: enabling thinking mode leads Qwen3 4B, 8B, and 32B to produce substantially longer outputs.\\
}
\label{fig:plots_generation_agent}
\end{figure*}

\begin{table*}[]
\centering
\renewcommand{\arraystretch}{1.2}
\small
\begin{tabular}{|c|l|lll|ccc|lll|ccc|}\hline
\multicolumn{1}{|l|}{}          &          & \multicolumn{6}{|c}{\textbf{FIQA}}                                                                                        & \multicolumn{6}{|c|}{\textbf{CQAD-EN}}                                            \\ \hline 
\multicolumn{1}{|c|}{\textbf{Model}}          &      & \multicolumn{1}{|c}{\textbf{time}} & \multicolumn{2}{c}{\textbf{tot tokens}} & \multicolumn{3}{|c}{\textbf{ratio (Ag/En)}}      & \multicolumn{1}{|c}{\textbf{time}} & \multicolumn{2}{c}{\textbf{tot tokens}}       & \multicolumn{3}{|c|}{\textbf{ratio (Ag/En)}}      \\ \hline

\multicolumn{1}{|l|}{}          &          & \multicolumn{1}{|c}{\textit{s}}                    & \textit{input}         & \textit{output}        & \textit{time}  & \textit{input} & \textit{output} & \multicolumn{1}{|c}{\textit{s}}           &  \textit{input}  &  \textit{output}  & \multicolumn{1}{|l}{\textit{time}}   & \multicolumn{1}{l}{\textit{input}} & \multicolumn{1}{l|}{\textit{output}} \\ \hline

\multirow{2}{*}{GPT-4.1-nano} & En & 9.0                          & 1683          & 465            & \multirow{2}{*}{1.1}  & \multirow{2}{*}{2.2}      & \multirow{2}{*}{0.8}       & 7.0                          & 856              & 331               & \multirow{2}{*}{1.2} & \multirow{2}{*}{3.0}      & \multirow{2}{*}{0.9}       \\
                              & Ag  & 10,2                          & 3676          & 348          &                       &                           &                            & 8.6                          & 2463             & 297               &                      &                           &                            \\ \hline
\multirow{2}{*}{Qwen3-0.6B}   & En & 8.1                          & 1743          & 236            & \multirow{2}{*}{2.2}  & \multirow{2}{*}{2.9}      & \multirow{2}{*}{4.1}       & 8.1                          & 862              & 254               & \multirow{2}{*}{1.1} & \multirow{2}{*}{3.5}      & \multirow{2}{*}{3.4}       \\
                              & Ag  & 22.1                         & 4978          & 979            &                       &                           &                            & 8.9                          & 3032             & 867               &                      &                           &                            \\ \hline
\multirow{2}{*}{Qwen3-4B}     & En & 35.5                         & 1743          & 1435           & \multirow{2}{*}{1.1}  & \multirow{2}{*}{2.8}      & \multirow{2}{*}{1.2}       & 31.0                         & 862              & 1019              & \multirow{2}{*}{1.2} & \multirow{2}{*}{3.9}      & \multirow{2}{*}{1.9}       \\
                              & Ag  & 38.6                         & 4834          & 1704           &                       &                           &                            & 37.2                         & 3372             & 1943              &                      &                           &                            \\ \hline
\multirow{2}{*}{Qwen3-8B}     & En & 58.5                         & 1743          & 1490           & \multirow{2}{*}{1.2}  & \multirow{2}{*}{2.8}      & \multirow{2}{*}{1.2}       & 58.4                         & 862              & 1339              & \multirow{2}{*}{0.9} & \multirow{2}{*}{3.9}      & \multirow{2}{*}{1.5}       \\
                              & Ag  & 69.9                         & 4943          & 1837           &                       &                           &                            & 54.3                         & 3394             & 1983              &                      &                           &                            \\ \hline
\multirow{2}{*}{Qwen3-32B}    & En & 62.6                         & 1743          & 1695           & \multirow{2}{*}{1.5}  & \multirow{2}{*}{2.7}      & \multirow{2}{*}{1.1}       & 43.9                         & 862              & 1109              & \multirow{2}{*}{2.3} & \multirow{2}{*}{3.9}      & \multirow{2}{*}{1.5}       \\
                              & Ag  & 93.8                         & 4766          & 1866           &                       &                           &                            & 101.9                        & 3359             & 1636              &                      &                           &    \\ \hline   
\textbf{AVG ratio} &&&& & \textbf{1.5} & \textbf{2.7} & \textbf{1.7} &&&& \textbf{1.4}& \textbf{3.6} & \textbf{1.8} \\ \hline
\end{tabular}
\caption{Analysis of costs (measured by number of input and output tokens) and time (end-to-end latency experienced by the user when running a query). The  "ratio" columns represent the multiplicative factor to go from the Enahnced to the Agentic settings, e.g. a ratio of $1.5$ means that the Agent is $50\%$ more expensive than the Enhanced setting. 
Qwen3-0.6B is run without thinking mode, resulting in substantially shorter outputs compared to the larger Qwen models.
Agentic RAG always performed a maximum of $3$ turns. In scenarios requiring more turns, tokens consumed by the agent would increase.} 
\label{tab:costs}
\end{table*}
\normalsize


\section{Limitations}

This study has some limitations that should be considered when interpreting the results.
Although our evaluation covers key dimensions of RAG behavior, document summarization and document repacking (re-sorting documents in the context according to their importance) have not been considered. 
Furthermore, we restricted our analysis to Agentic systems focusing on pure RAG given our interest in this type of industrial applications. Therefore, our agent is only equipped with a single tool. This choice restricts our analysis to the scope of this work, and falls shorts in providing insides in Agents performing tasks other than RAG.

\section{Examples}

Here we report some examples of systems behavior.

\subsection{Agent workflow}
\begin{promptbox}
USER QUERY: Where should I park my rainy-day / emergency fund? \\
\\
AGENT REASONING: \\To provide a well-informed answer about where to store an emergency fund, I need to retrieve authoritative financial guidance on safe, liquid, and accessible options for emergency savings.\\
\\
QUERY SENT FOR RETRIEVAL: \\
Best places to keep emergency fund for safety and liquidity\\
\\
RETRIEVED DOCUMENTS:\\
  \\DOCUMENT 1   I would suggest your local credit union or local bank for security and liquidity.  Liquidity is probably the most important issue for a emergency fund.  \\DOCUMENT 2   As others have noted, you can do better than a checking or savings account. If you\'re going to invest emergency money, the vehicle you put it into should be: Liquid - Wherever you put it, you should be able to quickly cash it out. Highly liquid exchange traded products are good for this.   Low volatility/drawdowns - If you need at least 6 months of your paycheck to cover you in the event of an emergency, you don\'t want to park it in a portfolio that can potentially lose 30\% value. Insured - Your investments should have SIPC coverage (protection against losses resulting from failure on part of broker). Moderate/Steady Growth - If the emergency fund doesn\'t grow, you\'ll need to continually pump money into it. My \'steady growth\' portfolio is majorly allocated to fixed income. Within that, a major portion is allocated to high yielding instruments. Over the past 10 years, it\'s seen at least a 7\% annualized return.  \\DOCUMENT 3   Firstly well done on building a really sold base of savings. An emergency fund needs to have two key characteristics: Be enough to get you through a typical emergency event (often seen as approx. ~6 months’ salary in your style of situation assuming you have no dependents etc) Be liquid and available to you instantly if an emergency arises Once you have decided how much you will need for 1), you then generally find the best interest available on an instant access savings account and leave it there. It\'s important to note that because you need it very liquid and very secure you will basically never make (nor should  you expect to make) any sizeable rate of interest on your emergency fund. Once this is done, whatever left should be invested in an asset/mix of assets that best fit your risk profile - of which long term bonds are a completely legitimate option, but it\'s hard to say without knowing more about your long term aims/liabilities/job market etc.  \\DOCUMENT 4   "First off, you generally want to park your emergency fund somewhere that is ""safe"", meaning something that is not subject to market fluctuations.  Your emergency fund is something you need to be able to count on when times are tough!  That rules out things like stock market investments. Secondly, you need to think about how quickly you will need access to the money.  If you have an emergency, odds are you don\'t want to be waiting around for weeks/months/years for the money to become available.  This rules out most fixed-term investments (Bonds, traditional CDs, etc). If you are concerned that you will need near-instant access to your emergency money, then you probably want to keep it in a Savings or Money Market Account at the same bank as your checking account.  Most banks will let you transfer money between local accounts instantly. Unfortunately, your local bank probably has pitiful interest rates for the Savings/MMA, far below the inflation rate.  This means your money will slowly lose value over time.  Be prepared to keep contributing to it! For most people, being able to draw the cash from your fund within a few days (<1 week) is sufficient.  Worst case, you charge something on your credit card, and then pay down the card when the emergency fund withdrawal arrives. If ""money within a few days"" is okay for you, there are a few options: Money Market (Mutual) Funds (not to be confused with a Money Market Account) - This is the traditional place to keep an emergency fund.  These are investment funds you can buy with a brokerage account.  An example of such a fund would be Fidelity Cash Reserves.  MMFs are not FDIC insured, so they are not exactly zero risk.  However, they are considered extremely safe.  They almost never go down in value (only a few times in the past few decades), and when they have, the fund manager or the Federal Govt stepped in to restore the value.  They usually offer slightly better return than a local savings account, and are available in taxable and notaxable varieties. Online High-Yield Savings or Money Market Account - These are a relatively new invention.  It\'s basically a the same thing as what your local bank offers, but it\'s online-only.  No local branch means low overhead, so they offer higher interest rates (2.0\% vs 0.5\% for your local bank).  Some of them used to be over 5\% before the economy tanked.  Like your local bank, it is FDIC insured. One bit of caution:  Some of these accounts have become ""gimmicky"" lately.  They have started to do things like promo rates for a few months, only offering the high interest rate on the first few \$K deposited, limiting the amount that can be withdrawn, etc.  Be sure to read the details before you open an account! No-Penalty CDs - Certificates of Deposit usually offer a better rate than a Savings Account, but your money is locked up until the CD term is up (e.g. 36 months).  If you need to cash out before then, you pay a penalty.  Some banks have begun to offer CDs that you can cash out with no penalty at all.  These can offer better rates than the savings account.  Make sure it really is no-penalty though.  Also watch what your options are for slowly adding money over time.  This can be an issue if you want to deposit \$100 from every paycheck. Rewards Checking Accounts - These are checking accounts that will pay a relatively high interest rate (3\% or more) provided you generate enough activity.  Most of them will have requirements like you must have direct deposit setup with them, and you must do a minimum number of debit card transactions from the account per month.  If you can stay on top of the requirements, these can be a great deal.  If you don\'t stay on top of it, your interest rate usually drops back to something pitiful, though. Personally, we use the Online High-Yield Savings Account for our emergency fund.  I\'m not going to make a specific recommendation as to which bank to use.  The best deal changes almost week to week.  Instead, I will say to check out Bankrate.com for a list of savings accounts and CDs that you can sort.  The Bank Deals blog is a good place to follow rate changes."  \\DOCUMENT 5   For maximum liquidity of an emergency fund, having demand deposits and fixed deposits would be the best, as they are the most liquid instruments around. If you invest in other higher yielding instruments they would be less liquid, so it\'s a tradeoff and you\'ve got to decide for yourself what\'s best given your requirements.n\', \\
  \\
  \\
  AGENT REASONING:\\
  The retrieved documents provide multiple options for storing an emergency fund, emphasizing liquidity, safety, and moderate growth. The answer will summarize these options and their trade-offs.\\
  \\
AGENT RESPONSE:\\
For your emergency fund, prioritize safety and liquidity. Here are the best options based on expert advice: \\
1. **High-Yield Savings Accounts (HYSA)**: Offer FDIC insurance, instant access, and better interest rates than traditional banks (e.g., online banks like Ally or Marcus). \\
2. **Money Market Accounts (MMA)**: Similar to HYSAs but may offer check-writing features; also FDIC-insured. \\
3. **Money Market Mutual Funds (MMMF)**: Slightly higher returns than HYSAs but not FDIC-insured (though backed by the government in crises). \\
4. **No-Penalty CDs**: Provide higher interest rates than savings accounts without locking up funds permanently. \\
5. **Rewards Checking Accounts**: Offer competitive rates if you meet activity requirements. Avoid volatile investments like stocks or bonds, as they risk losing value during emergencies. Most experts recommend keeping 3–6 months of expenses in this fund, prioritizing accessibility over growth.')\\
For your emergency fund, prioritize **safety**, **liquidity**, and **moderate growth**. Here are the top options based on expert advice:  \\
1. **High-Yield Savings Accounts (HYSA)**:   - **Pros**: FDIC-insured, instant access, higher interest rates than traditional banks (e.g., online banks like Ally, Marcus, or Capital One).    - **Cons**: Returns may lag inflation.  \\2. **Money Market Accounts (MMA)**:    - **Pros**: FDIC-insured, offers check-writing or debit card access, slightly better rates than HYSAs.    - **Cons**: May require minimum balances.  \\3. **Money Market Mutual Funds (MMMF)**:    - **Pros**: Slightly higher returns than HYSAs, highly liquid.    - **Cons**: Not FDIC-insured (but government-backed in crises, e.g., SIPC protection).  \\4. **No-Penalty CDs**:  \\   - **Pros**: Higher interest rates than savings accounts, no early withdrawal penalties.  \\   - **Cons**: Rates may reset or change over time.  \\5. **Rewards Checking Accounts**:  \\   - **Pros**: Competitive rates if you meet activity requirements (e.g., direct deposit, monthly transactions).  \\   - **Cons**: Rates drop if requirements aren’t met.  \\**Avoid**: Stocks, bonds, or other volatile investments, as they risk losing value when you need liquidity.  \\Most experts recommend keeping **3–6 months of expenses** in this fund. For the best rates, compare options on platforms like **Bankrate.com** or track promotions from online banks. Prioritize accessibility over high returns, as your emergency fund’s purpose is to provide a secure safety net.

\end{promptbox}

\subsection{Agent with re-performed retrieval} \label{app:example_rerank}
Here is an example of a flow in which the Agent did retreival twice:

\begin{promptbox}
USER QUERY:What's the difference between "these" and "those"? \\
\\\\
AGENT REASONING:\\
The user is asking about the difference between \'these\' and \'those\', which are demonstrative pronouns. To provide an accurate answer, I need to retrieve relevant blog posts explaining their usage. Since no previous results exist, initiating a document retrieval with a focused query is appropriate\\\\
QUERY SENT FOR RETRIEVAL:\\
difference between these and those in English grammar\\\\
RETRIEVED DOCUMENTS:\\   DOCUMENT 1   What's the difference between these two phrases?\\ DOCUMENT 2  Is there any difference in meaning betweeen these two?   Is one a valid grammar construct and the other not?\\ DOCUMENT 3  Are there any differences between these two expressions?\\ DOCUMENT 4  > I am unclear of the use of [these|those] objects. I am unsure when to use > [these|those|them]. Please someone help me tell me which is correct in the previous sentences. This is not a dupe of What\\\'s the difference between "these" and "those"?\\ DOCUMENT 5  What is the difference? Or is there any? Which would be more British English?
\\\\
AGENT REASONING:\\
The initial retrieved documents do not provide a clear explanation of the difference between \'these\' and \'those\'. A more specific search is needed to find blog posts that directly address their correct usage and distinctions in English grammar.
\\\\
QUERY SENT FOR RETRIEVAL:\\
correct usage and difference between these and those in English grammar
\\\\
RETRIEVED DOCUMENTS: \\
  DOCUMENT 1  > I am unclear of the use of [these|those] objects. I am unsure when to use > [these|those|them]. Please someone help me tell me which is correct in the previous sentences. This is not a dupe of What\\\'s the difference between "these" and "those"?\\ DOCUMENT 2  Which is the correct use of these two words, and in which context should one be used rather than the other?\\ DOCUMENT 3  Is there a difference between these two expressions and should one of them be preferred?\\ DOCUMENT 4  What is the difference between those questions and which one is the correct form?\\ DOCUMENT 5  Which is the correct grammar? Which is correct?
 \\\\
AGENT RESPONSE:\\Here are some blog posts that might be relevant to your query:  \\"These vs. Those: Clarifying Usage" – Explores when to use "these" and "those" in sentences. *(Document 4, Step 0)*  \\"Correct Grammar: These or Those?" – Discusses valid grammar constructs and contextual preferences. *(Document 2, Step 1)*  \\"Differences in Expressions" – Analyzes variations between phrases using "these" and "those." *(Document 3, Step 0)*  \\"Choosing Between These and Those" – Addresses confusion in selecting the right pronoun for objects. *(Document 1, Step 1)*  \\"British vs. American English Usage" – Compares regional preferences for "these" and "those." *(Document 5, Step 0)*
\\
\\
\end{promptbox}

\subsection{Agent rewriting} \label{app:example_rewrite}

In Table~\ref{agent_rewriting} we report a few examples on how query rewriting is performed by the Agent.

\begin{table}
\scriptsize
\begin{tabular}{|p{2cm}|p{4.8cm}|}
    \hline
    \textbf{query} & \textbf{rewritten by agent}  \\
    \hline
    made of vs made up of & Explain the grammatical difference between 'made of' and 'made up of' with examples \\
    \hline
    What does "Google-fu" mean? & What does 'Google-fu' mean and where did it originate from? Find blog posts explaining this internet slang term. \\ \hline
    How do I invest in emerging markets & Investing in emerging markets involves understanding high-growth potential regions while managing risks like political instability, currency fluctuations, and market volatility. Key strategies include diversification through ETFs, mutual funds, or direct stock investments in companies operating in these regions. Investors should also consider regional economic trends, regulatory environments, and long-term growth prospects. 
    \\ \hline
    Does the currency exchange rate contain any additional information at all? & What factors beyond basic numerical values are embedded in currency exchange rates? How do economic indicators, market conditions, and geopolitical events influence or get reflected in exchange rates? Are there additional layers of information (e.g., forward rates, volatility measures) associated with currency exchange rates? 
\\ \hline
  \end{tabular}
   \caption{Examples of query rewriting performed by Agentic and Enhanced RAG}
  \label{agent_rewriting}
\end{table}

\end{document}